\documentclass[letterpaper]{article} % DO NOT CHANGE THIS
\usepackage{aaai25}  % DO NOT CHANGE THIS
\usepackage{times}  % DO NOT CHANGE THIS
\usepackage{helvet}  % DO NOT CHANGE THIS
\usepackage{courier}  % DO NOT CHANGE THIS
\usepackage[hyphens]{url}  % DO NOT CHANGE THIS
\usepackage{graphicx} % DO NOT CHANGE THIS
\usepackage{natbib}  % DO NOT CHANGE THIS AND DO NOT ADD ANY OPTIONS TO IT
\usepackage{caption} % DO NOT CHANGE THIS AND DO NOT ADD ANY OPTIONS TO IT
\usepackage{algorithm}
\usepackage{algorithmic}
\usepackage{multirow}

\usepackage{newfloat}
\usepackage{listings}
\DeclareCaptionStyle{ruled}{labelfont=normalfont,labelsep=colon,strut=off} % DO NOT CHANGE THIS
\floatstyle{ruled}
\newfloat{listing}{tb}{lst}{}
\floatname{listing}{Listing}
\usepackage{macros}
\usepackage{pgfplots}
\usepackage{enumitem}
\usepackage{subcaption}

\newtheorem{definition}{Definition}

\usepackage[textsize=tiny,
	 disable,
	]{todonotes}

\newcommand{\ben}[2][]{\todo[color={green!20},linecolor={green!100},#1,size=\tiny]{#2}}

\title{Goal Recognition using Actor-Critic Optimization}
\author {
    Ben Nageris\textsuperscript{\rm 1},
    Felipe Meneguzzi\textsuperscript{\rm 2},
    Reuth Mirsky\textsuperscript{\rm 1, 3}
}
\affiliations {
    \textsuperscript{\rm 1}Bar Ilan University, Israel \\
    \textsuperscript{\rm 2}University of Aberdeen, Scotland, UK \\
	\textsuperscript{\rm 3}Tufts University, MA, USA \\
    benabraham.nageris@live.biu.ac.il, felipe.meneguzzi@abdn.ac.uk, reuth.mirsky@tufts.edu
}

\usepackage{bibentry}
\begin{document}

\maketitle

%%% Use this environment to include an abstract of your paper.

\begin{abstract}
    % !TEX root = ../main.tex
\ben{Remove "optimization" from the title?}
Goal Recognition aims to infer an agent's goal from a sequence of observations. 
Existing approaches often rely on manually engineered domains and discrete representations. Deep Recognition using Actor-Critic Optimization (DRACO) is a novel approach based on deep reinforcement learning that overcomes these limitations by providing two key contributions. 
First, it is the first goal recognition algorithm that learns a set of policy networks from unstructured data and uses them for inference.
Second, DRACO introduces new metrics for assessing goal hypotheses through continuous policy representations. 
DRACO achieves state-of-the-art performance for goal recognition in discrete settings while not using the structured inputs used by existing approaches. Moreover, it outperforms these approaches in more challenging, continuous settings at substantially reduced costs in both computing and memory. 
Together, these results showcase the robustness of the new algorithm, bridging traditional goal recognition and deep reinforcement learning.

%% FRM - Previous version of the abstract
% Goal Recognition is the task of inferring the goal of an agent from a sequence of observations, often represented as a series of predefined symbols.
% However, realistic observations can consist of unstructured data such as videos or raw text, rather than a set of symbols.  
% To bridge this gap between the real world and the input of goal recognition algorithms, many approaches rely on manually crafted rules to process raw data, which leads to brittleness and lack of scalability. 
% We address this limitation by extending the traditional goal recognition problem to directly handle raw, continuous data instead of manually crafted symbols and rules. 
% We then develop a new algorithm that automatically acquires a domain model using deep reinforcement learning.
% \textbf{The algorithm, Deep Recognition using Actor-Critic Optimization (DRACO), performs end-to-end goal recognition from raw data.}
% Empirical results show that DRACO reaches similar results to the state-of-the-art GR in discrete domains while using only the image-based inputs, and outperforms it when transitioning to more challenging, continuous domains. These results demonstrate the robustness of the new algorithm, that bridges between traditional goal recognition and deep reinforcement learning.
\end{abstract}

%%%%%%%%%%%%%%%%%%%%%%%%%%%%%%%%%%%%%%%%%%%%%%%%%%%%%%%%%%%%%%%%%%%%%%%%

\section{Introduction}\label{sec:intro}
    % !TEX root = ../main.tex
Goal recognition (GR) is a fundamental problem in AI where the objective is to infer the terminal goal of an observed agent from a sequence of observations \cite{sukthankar2014plan}. %,mirsky2021introduction}. %% FRM Since these are citations to the definition of the problem itself, it's better to cite surveys and more basic material. I just happen to be ours :-D. %%  RM: We can definitely add them once the paper is accepted, but writing down 2/3 of the authors' names in the first sentence is a bit too much against anonymization.
%\cite{chiari2022goal,pattison2011accurately}. %, meneguzzi2021survey, mirsky2021introduction}. 
GR techniques can be used for estimating people's paths~\cite{vered2017online}, recognizing actions in a kitchen for a service robot to fetch the required ingredients~\cite{kautz1986generalized,shvo2022proactive}, or even alerting an operator when a client executes a suspicious sequence of actions in e-commerce software~\cite{qin2004attack}. 
% Figure~\ref{fig:comparison} (top) illustrates the common pipeline for solving such GR tasks. 
A recognition task contrasts a sequence of observations, represented as discrete, symbolic actions, with a domain theory that describes the domain's possible actions to infer the likely goal from the observations. 
This process inherits three fundamental limitations of using a symbolic domain theory.
First, using such domain theories forces the recognizer to rely on accurate and relevant symbols and limits the size of feasible problems. 
Second, these underlying models often represent an optimal plan for a goal but require additional extensions to handle noise in the recognition process. 
Third, recognizing goals in continuous domains necessitates a discretization process that may jeopardize recognition~\cite{kaminka2018plan}. 

% \begin{figure}
%      \centering
%      \begin{subfigure}[b]{0.5\textwidth}
%          \centering
%          \includegraphics[width=0.9\textwidth]{Figures/comparison.pdf}
%          \label{fig:GR}
%      \end{subfigure}

%         \caption{
%         Traditional activity and goal recognition pipeline (top) and our combined approach (bottom). The agent and the potential goals are marked with a red triangle and green squares accordingly.
%         % is marked with a red triangle and the potential goals are marked using green squares. 
%         Each pyramid expresses a distinct recognition process: goal, plan, or activity.} %obviates an intermediate symbolic representation.}
%         \label{fig:comparison}
% \end{figure}

\begin{figure}[t]
    \begin{center}
        \includegraphics[width=0.45\textwidth]{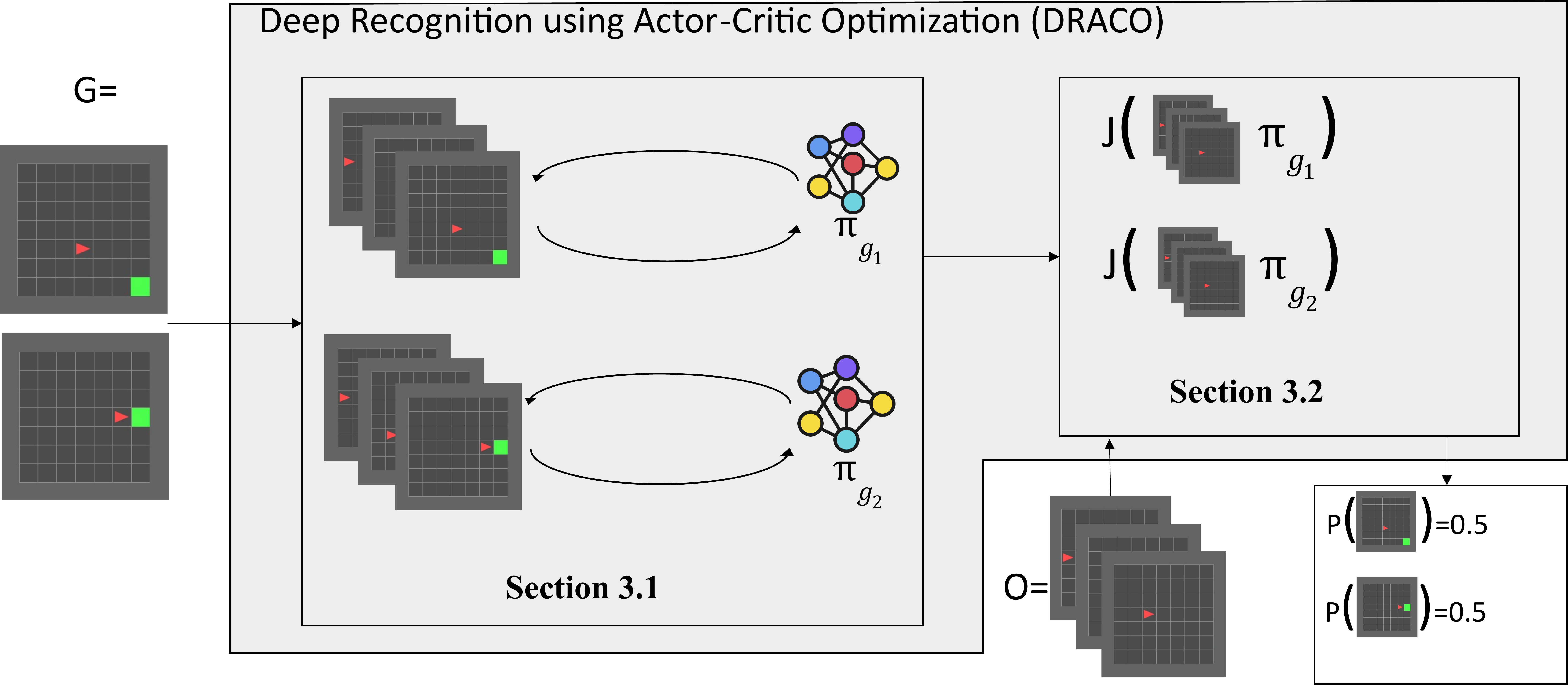}
        \caption{Overview of DRACO.}
    \label{fig:draco_overview}
    \label{fig:draco}
    \end{center}
\end{figure}

%% FRM - The paragraph below did not have a good connection to what we originally stated above. 
% The most straightforward solution is to discretize the environment and map each continuous state into a discrete state. 
% However, choosing a suitable granularity for discretization imposes a non-trivial tradeoff. 
% Finer granularity may result in more accurate outcomes but at the price of significantly higher computation costs. 
% By contrast, a coarser granularity may cause information loss by mapping similar but meaningfully different continuous states into the same discrete state. 
% Previous work shows that for any discretization granularity chosen in advance, there is a goal recognition problem that remains unsolvable \cite{kaminka2018plan}. 

This paper develops the \textit{Deep Recognition using Actor-Critic Optimization (DRACO)} algorithm to overcome these limitations. %, as illustrated in Figure~\ref{fig:comparison} (bottom). 
DRACO's input is vector-based raw data, and it outputs a distribution over the goals the observed agent might be pursuing. 
DRACO offers two key contributions over existing approaches to GR. 
First, DRACO uses interactions with the environment to learn goal-dependent neural networks (NNs), representing parameterized policies that allow us to estimate the likelihood of the observed agent pursuing each goal in virtually any type of learnable domain. Such domains include continuous domains, arbitrarily large discrete domains, and domains whose observations consist entirely of raw data. DRACO's NNs replace the need to run costly planners in real-time, and their policies can potentially be transferred to represent new goals using transfer learning \cite{shamirodgr,taylor2009transfer}.
These learned policies obviate the requirement of carefully engineered domain theories and instead rely on data of the same nature as the observations we expect at recognition time. 
Second, this paper develops two distance measures to compare observations with the learned goal-conditioned policies, derived, respectively from Wasserstein distance, the statistical Z-score metric. 

The empirical evaluation consists of two very different domains: \textit{MiniGrid} supports discrete and continuous state space representation and a discrete action representation, while \textit{Panda-Gym} consists of continuous state and action spaces. 
Each domain was evaluated on various scenarios, varied by difficulty of goal disambiguation, observability levels, and noise level. 
Our comparison of DRACO with leading algorithms both in discrete and continuous environments shows that it outperforms state-of-the-art in both types of environment. 
In discrete domains DRACO does so even when given fewer structured inputs and fewer training episodes, whereas in continuous environments it dramatically outperforms previous approaches. 
Our secondary contribution is test bed for domains with continuous state and action spaces.
It leverages existing RL simulations into an evaluation domain with automated instance generation. 
\ben{The second contribution is a testbed to perform goal recognition which both state-of-the-art and other approaches can interpret}

% \ben{I changed to a paragraph}To conclude, the contributions of this paper are fourfold:
% First, a robust goal recognition algorithm for both discrete and continuous domain.
% Second, two evaluation metrics for inferring the distance of an observation sequence from a policy.
% Third, testbed for goal recognition problems with a problem generator.
% Forth, empirical evaluation spanning a large variery of domain and instance features.
% \ben{Maybe change to a paragraph?}\begin{enumerate}
%     \item A robust goal recognition algorithm for both discrete and continuous domains.
%     \item Two evaluation metrics for inferring the distance of an observation sequence from a policy.
%     \item A testbed for goal recognition problems with a problem generator.
%     \item An empirical evaluation spanning a large variery of domain and instance features.
% \end{enumerate}
% \frm[inline]{We ought to trim the intro down, let's consider removing the Figure.}

\section{Background}\label{sec:background}
    % !TEX root = ../main.tex
% Previous work \cite{amado2022goal} develops a framework to perform GR as Q-Learning (GRAQL).
% This framework consists of two main stages: the learning and the inference stage.
% The learning stage uses the off-the-shelf Q-Learning algorithm to learn a set of Q-functions for each possible goal separately.
% In addition, the inferring stage is responsible for recognizing the goal of an actor given a sequence of observations.
% For each potential goal, it measures the distance, using standard machine learning metrics (KL divergence, MaxUtil and Divergence Point), to the observation and chooses the one that minimizes this value.

 % \begin{definition}
 % \label{def:GR}
 % \textbf{Goal Recognition}
 \subsection{Recognition Problems}
 Goal and Activity recognition are related but distinct tasks~\cite{van2021activity,meneguzzi2021survey,mirsky2021introduction,sukthankar2014plan}, which have three main components. 
 First, the \textit{environment} specifies the problem's dynamics and settings. 
 Second, the \textit{observed agent} represents the agent that acts in the environment and pursues some objective. 
 We note that often, this \textit{observed agent} is called an \textit{actor}, but we use \textit{observed agent} instead to differentiate it from the Actor in the Actor-Critic algorithm description.
 Third, the \textit{observer} watches the observed agent as it acts in the environment. 
 Beyond those similarities, each recognition problem has its own properties. 
 \textit{Activity Recognition} (AR) problems focus on labeling a sequence of possibly noisy sensor inputs with a single usually unstructured label, denoting a low-level activity of the observed agent, while
 % \textit{Plan Recognition} focuses on finding a specific structured plan (set of actions) that is being followed by an agent, including the goal being pursued. 
 \textit{Goal Recognition} (GR) is the most abstract level of inference focusing on identifying the highest-level goal or overarching objective of the agent's actions. 
 The problem formulation in this work provides a synthesis of AR and GR, as it requires a sequence of low-level sensor data as input and outputs a prediction about the goal of the observed agent. %This variant of a recognition problem is sometimes called \textit{Intention Recognition (IR)} or \textit{Objective Recognition}. As most recognition algorithms tackle either the GR or the IR problems, we focus the related work on them.
We use a similar formulation for goal recognition as planning (\citet{ramirez2009plan}):
\begin{definition}
\label{def:GR}
    A \textit{Goal Recognition (GR)} problem is a tuple $\langle \planningdomain, \goals, \observations \rangle$, such that $\planningdomain$ is domain theory, $\goals$ is an exhaustive and mutually exclusive set of potential goals, and $\observations$ is a set of observations. A solution for a GR problem is a goal $g \in \goals$ that $\observations$ explains according to $\planningdomain$.
\end{definition}
The domain theory $\planningdomain$ is a description of the environment. It can be prescribed in various ways, including plan graphs \cite{avrahami2005fast,e2015fast}, a planning domain \cite{ramirez2009plan,masters2019cost,pereira2020landmark}, or a Markov Decision Process (MDP) \cite{ramirez2011goal,amado2022goal}.
As we combine GR and AR, such that both the goal and the observations can be either a symbolic state representation (e.g., coordinates in space), or other raw data such as images or floating point numbers vectors. 
% Each goal $g$ can be expressed as a set of labels over possible goal states, or a probability distribution over alternative sets of these states.\frm{Not sure this sentence makes sense here.} 
Each goal $g$ can be expressed as a formula, a set of labels over possible goal states, or a reward function over alternative sets of these states.  
Similarly, the observations $\observations$ can be either a sequence of symbolic states or raw data. 
Finally, the observer can perceive all actions and state sets or only part of them.

% In general data-driven-based approaches enjoy smoother adjustment when facing a new problem compared to the symbolic approach as they redundant the use of a specific planning or parsing library to the exact problem.
% Although presenting a severe flexibility improvement, not all data-driven approaches have the same level of robustness. 
% DL approaches require extensive pre-collected data in the learning phase. Nevertheless, RL approaches have better flexibility because they first collect experiences (state, action, and reward on-the-fly data) in the learning phase via environment exploration and then exploit this experience to better react in the environment.
% In both data-driven approaches, the domain expert specifies less information about the framework compared to the symbolic approach. While existing data-driven approaches have many advantages, they only work on discrete domains.

% \subsubsection{Reinforcement Learning}
% % Reinforcement Learning (RL) is an area of machine learning concerned with how intelligent agents should take actions in an environment to maximize the notion of cumulative reward.  
% %% FRM the sentence above is incorrect, there is RL that tries to maximize average reward, or mininimize regret.
% (RL) is an area of machine learning concerned with how intelligent agents should take action in an environment that provides feedback in the form of a reward signal.  
% % The main advantage of policy-based algorithms is that they usually convergence faster due to having less variance in the action space, while value-based functions are simpler.
% 

\subsection{Planning Domain Definition Language (PDDL)}
\ben{Added now, feel free to modify}
% Symbolic GR approaches relies on having a complete domain representation, parsing, and inferring in realtime. A common format for describing a planning problem is \textit{PDDL}.

% \begin{definition}
\label{def:PDDL}
PDDL is a formal language for describing planning problems, and introduced to standardize the representation of problems in automated planning and to serve as a common standard for benchmarking planning algorithms. 
A PDDL problem description typically consists of a domain file and a problem file. The domain file specifies the predicates, types, constants, and operators (actions) available, while the problem file defines the specific objects, initial state, and goal conditions.
% \end{definition}

Symbolic GR approaches relies on having a deliberate domain representation (\textit{PDDL} for example) which is susceptible to noise, and require complete and precise specification usually made by a domain expert manually (\cite{ramirez2009plan, RamirezGeffner2010}).
This representation makes them often difficult to scale and costly in terms of real time computation.
% Crafting this domain representation is challenging and requires an expert in the domain to describe a model.
% Standardizing the planning problem's input made it possible to execute generic planners. However, creafting this domain description remains the main issue in using 

\subsection{Reinforcement Learning}
\ben{Maybe delete this part? RLC is an RL conference, I think we should change this part to explain maybe PDDL}
This work focuses on Reinforcement Learning (RL)-based recognition. A commonly used domain model in RL is MDP \cite{sutton2018reinforcement}.
MDP is used to model decision-making processes, and can encapsulate both discrete and continuous state and action spaces.
% is used in RL that is used to model decision-making, and can encapsulate both discrete and continuous state and action spaces is an \textit{Markov Decision Process} (MDP) \cite{sutton2018reinforcement}.

\begin{definition}
\label{def:MDP}
An \textbf{MDP} M, is a 4-tuple $\tuple{\states, \actions, P, R}$ such that $\states$ are the states in the environment, $\actions$ is the set of actions the agent can execute, $P$ is a transition function, and $\rewards$ is a reward function.
A transition function $\prob{s' \mid s, a}$ returns the probability of transitioning from state $s$ to state $s'$ after taking action $a$, and a reward function $R(s, a, s')$ returns the reward an agent obtains when it transitions from $s$ to $s'$ using action $a$.
\end{definition}
The solution to an MDP is a \textit{policy} $\pi$, which in RL often consists of a stochastic function $\policy(a \mid s)$ that defines the probability of the agent taking action $a \in \actions$ in state $s \in \states$. 
A policy $\optimalpolicy$ is optimal if it maximizes the expected return of all states of the MDP. 
In Deep RL, NNs estimate these policies. 
In DRACO, learned policies represent the potential behavior of an observed agent rather than the desired behavior of the learner itself, and hence we can shape the reward differently for learning each goal-dependent policy. 

\section{DRACO}\label{sec:draco}
DRACO relies on the 
% RL 
assumption that the environment behaves as an MDP (Definition \ref{def:MDP}) and in practice, we only need to represent states and actions as our domain theory $\planningdomain$ from Definition \ref{def:GR}:
\begin{definition}
%     A GR as RL problem is a tuple $\tuple{\planningdomain,G,O}$:
% \begin{itemize}[nosep]
%     \item $\planningdomain=\tuple{\states, \actions}$ is a set of $n$-dimensional states and $m$-dimensional actions, both of which can be discrete or continuous, $\states \subseteq \Real^{n}$ and $\actions \subseteq \Real^{m}$ %\reuth{n,m are the dimensions of the state/action spaces}; 
%     \item $\goals \subseteq \states$ is a set of potential goals, out of which we assume the observed agent pursues exactly one; and
%     \item $\observations = s_0, a_0, s_1, a_1, s_2, a_2, \ldots$ is a sequence of state-action observations, such that $s_i \in S$ and $a_i \in A$.
% \end{itemize}
A GR as RL problem is a tuple $\tuple{\planningdomain,G,O}$ such that $\planningdomain=\tuple{\states, \actions}$ is a set of $n$-dimensional states and $m$-dimensional actions, both of which can be discrete or continuous, $\states \subseteq \Real^{n}$ and $\actions \subseteq \Real^{m}$; $\goals \subseteq \states$ is a set of potential goals, out of which we assume the observed agent pursues exactly one; and \ben{ Change observation definition} $\observations = s_0, a_0, s_1, a_1, s_2, a_2, \ldots$ is a sequence of state-action observations, such that $s_i \in S$ and $a_i \in A$.
\end{definition} 
This definition is purposefully broad, allowing for different possible instantiations of the problem: States can be images or continuous data such as a robot's joints' location and velocities. This representation obviates the need to manually prescribe labels or symbols to specific states. %with actions being similarly broad. 
While we do not assume explicit transition dynamics to be part of the problem, we assume having access to the environment or some model of the environment, such as a sampling model or simulator. These assumptions free us from explicitly prescribing action dynamics as part of a recognition problem and position our solution as a model-free RL technique. 

% As a goal recognition algorithm, DRACO is required to reason about which goal is most likely to explain a set of observations. It does so by comparing this sequence to policies, such that each policy represents one possible goal from $G$. 

Figure~\ref{fig:draco} illustrates the recognition process in DRACO. %, which we explain in this section. %% FRM second sentence is redundant.
It starts by receiving a set of goals ($|\goals|=2$ in the figure), and the observation sequence $O$ ($|O|=3$ in the figure). 
To solve the GR problem, we first discuss how DRACO learns a set of policies, one policy for each goal, using simulated interactions. 
Then, we explain how DRACO compares the observation sequence $O$ against each goal-oriented policy.
% to estimate the likelihood of the observed agent producing $O$ given each policy. 
Key to this comparison is the metric for processing observations and comparing them to policies. 
In this work, we propose \ben{three? do we keep the state only metric?}{two} different metrics. 
% While previous work makes strong assumptions of discreteness, we extend these metrics to continuous domains in Section~\ref{sec:metrics}. 
The algorithm then outputs a distribution over the set of goals to represent how likely each goal to explain the observation.
\subsection{Policy Learning} 
\label{sec:learn}

GR often decouples the description of the problem, including the set of recognizable goals, from a specific observation sequence \cite{ramirez2011goal,masters2019cost}. 
There is a clear gain in precomputing these policies, as the learning process is the costliest component of the pipeline, and we can assume prior knowledge of the goals in realistic domains.  %add Meneguzzi's survey if accepted 
Consistent with this, DRACO learns a set of policies offline, one policy $\pi_g$ for each goal $g \in G$. 
% Note that a significant strength of this work is that it does not require the learned policies to be an accurate\frm{Accurate or optimal?} representation of actors following each goal. 
% Rather, they only need to be accurate enough such that a policy for goal $g_i$ is more similar to the behavior of an actor pursuing $g_i$ than to the behavior of an actor pursuing any other goal.
A significant strength of this approach is that it does not require the learned policies to represent agents pursuing each goal optimally. Rather, they only need to hold that a policy for goal $g_i$ is more similar to the agent's behavior of pursuing $g_i$ than to the agent's behavior of pursuing any other goal.

DRACO learns goal-dependent policies $\pi_g$ using Actor-Critic policy gradient methods. 
These algorithms naturally support continuous state and action spaces, which are crucial in many physical environments, such as robotics, trajectory recognition, etc. 
We use the sampling model of the environment to compute the policy $\policy_{g}$ for each candidate goal $g$. % to generate a policy targeting each goal. % in Lines~\ref{alg:learnStart}--\ref{alg:learnEnd}. 
We assume our model can simulate environmental dynamics but does not contain explicit reward information. 
This assumption is consistent with most real-world environments, such as movement in robotic arms, which do not include explicit reward specifications. 
%\reuth{Revised, but I think we need to clarify what we mean by "real world environmnets do not include explicit description of reward" -- which real world domains are these? Simulators or "the real world"?}
% The transitions are determined according to actual interactions with the environment, and the reward function for goal $g, R_g$, is set to give a reward for reaching $g$, thus motivating the agent to move towards this state (Line \ref{alg:setReward}). 
%For each goal,  we reset its dynamics to a predefined initial state, and provide a domain-dependent reward function. \frm{Ok, we need to discuss this motivation thing.}
Given this model, for each goal $g$, DRACO generates a different reward function, meant to shape the agent's preferences to reach that goal.
% Given this model, DRACO generates a different reward function per goal, meant to shape the agent's preferences to reach that goal.
% \ben{We need to verify in Rg definition we note that its only the L1 distance}
% For goal $g$, the reward function returns a positive value when the agent reaches $g$. 
% and a negative reward otherwise. %The return is thus accumulated via backpropagation. %Formally,

% \begin{equation}
%     R_g(s,a) = 
%     \begin{cases}
%     +C, & \text{if } env(s, a) = g \\
%     -1,   & \text{otherwise}
%     \end{cases}
% \end{equation}

%Second, the reward gives incentives throughout the path to the goal helping the algorithm to shape its decisions towards the correct actions in its way to the terminal state. \reuth{I'm not sure I follow this part, and why we need to state it here. Is it something that is general across all domains in which we will deploy DRACO?}
The reward function has a large influence on the learning part of the framework. 
Sometimes, the sparsity of the positive reward might require us to shape it further to help the agent learn how to reach the goal. 
Several goal-conditioned RL algorithms apply here \cite{andrychowicz2017hindsight,durugkar2021adversarial,kaelbling1993learning,schaul2015universal}, and transfer learning techniques can be used to learn new policies for new goals quickly, thus enabling DRACO to scale quickly with the number of goals. However, to minimize variability in this work, in our empirical evaluation we replace these sophisticated algorithms with a simple aggregated reward. 
To train the goal-dependent policy $\pi_g$, we consider the reward for action $a$ in state $s$ to be:
% , $R_g(s,a)$ as the $L_1$ distance between goal $g$ and the state after the action $a(s)$. 
% To train the goal-dependent policy $\pi_g$, we compute the reward for action $a$ in state $s$, $R_g(s,a)$ as the $L_1$ distance between goal $g$ and the state after the action $a(s)$. 
%
\begin{equation}
\label{eq:aggregated_reward}
    R_g(s,a) = (-1) * \|  a(s) - g \|_{L_1}
    \normalsize
\end{equation}
% With this reward function $R_g$ at hand, DRACO can learn the corresponding policy for each goal $g$. 

% 
% \reuth{I am tempted to leave this revised part about the reward here, because we're already at the end of the 3th page and we haven't presented a single equation/algorithm yet... Just for visibility reasons.}
% \frm{Agreed}
% 
%  since motivation is sometimes required to minimize the learning time.
% Specifically, in this paper, we used the environment's default reward function.
% In Minigrid, we used sparse reward function  \reuth{Ben will update this according to the continuous reward function.}\frm{We might want to move some of the discussion about specific environments to Section~\ref{sec:results}}
% 
% 
% Nevertheless, in panda-gym,\frm{Again, any discussion of domain-specific stuff, we move to \ref{sec:results}} we use the dense reward function which in which the reward is the summed difference between the goal and the state in all axes
% 
% 
% 
% This MDP is then used as the underlying environment for an RL problem (Line \ref{alg:callPPO}). 
% By doing so, the challenge of acquiring a domain theory becomes an RL problem with its respective decisions: selecting the most appropriate algorithm for learning and tuning its hyperparameters. 
% In this work, we use PPO and SAC to learn such policies, as they can efficiently learn policies both in continuous and discrete domains, so they do not limit our state representation. 
% 
% 
%The agent explores the environment separately for each possible goal and learns a policy to reach the goal. For example, 
\noindent \textbf{Learning Architectures}
\label{sec:nn}
This work focuses on Actor-Critic (AC) approaches to learn the goal-dependent policies since they are the leading approaches in policy-based continuous RL.
Each AC algorithm requires two NNs per goal (actor NN and critic NN), so its total number of trained NNs is $2 \times |\goals|$. 
% The main difference between PPO and SAC is that SAC cannot handle a discrete action space.
The Actor network outputs a probability of taking each action, the size of its output is the number of valid actions. By contrast, the critic network, which outputs the expected value for being in this particular state, has a single neuron in its last layer.
Once DRACO computes these two NNs, it uses the Actor's output to generate policies for each goal ($\policy_g$) and an observation sequence $O$ for GR: inferring the goal ($g \in G$) that $O$ explains. 
\ben{Change the network usage for each metric - in state-only observations we use the critic network}The reason that we chose to use the Actor network's output rather than the Critic network's is that the Actor network outputs an action rather than a value, which is easier to compare to the observation sequence.
\subsection{Likelihood Estimation of Observations (inference)}
\label{sec:infer}

Following the Bayesian formulation from \citet{RamirezGeffner2010}, we compute the probability $\prob{g \mid O}$ of a goal $g \in \goals$ conditioned on observations $O$ by computing the probability $\prob{O \mid g}$ of the observations conditioned on the goals. 
%
%Since we train one policy $\policy_{g}$ for each goal $g \in \goals$, we can see such learned policies as a function that provides the probability of an agent choosing actions in an environment, conditioned by the goal this agent is pursuing, so that $\policy_{g}(a \mid s) \defeq \prob{a \mid s, g}$. 
%
% Now, recall that the softmax policies learned by Actor-Critic algorithms encode, for each state in the environment, the probability with which an approximately optimal agent chooses an action towards a goal within such environment. 
The key challenge here is to estimate the likelihood of $O$ given $g$, which is the likelihood of taking the action $a$ when in state $s$ according to $\policy_{g}$. 
Formally, $\policy_{g}(a \mid s) \defeq \prob{a \mid s, g}$.
For this estimation, we need to design a distance function between an observation sequence of the form $s_0, a_0, s_1, a_1, \ldots$ and a policy $\policy_g(a_i \mid s_i)$. 
This function should be able to handle continuous spaces to enable goal recognition in such environments. %, unlike previous work, which makes strong assumptions about the nature of the state and goal spaces~\cite{amado2022goal}.  
% Thus, in order to compute the probability of a goal, conditioned on a sequence of observations, 
We can then compute an aggregate distance $\distance$ of the observations $O$ to the observations we expect an agent to generate if it follows a policy towards that goal at each step in the observations. 
This metric allows us to compute the probability of the observations for each goal using the softmin of such distances:
\begin{equation}
        \prob{O \mid g} = \softmin_{g \in \goals}(\distance(O,\policy_{g})) \label{eq:softmin}
\end{equation}
Then, using the assumption of exhaustive and mutually exclusive goal hypotheses from Definition~\ref{def:GR}, we can compute the probability of each goal hypothesis by normalizing over the individually computed conditional probabilities in Eq.~\ref{eq:ramirez}. 
\begin{equation}
% \end{align}
% \begin{align}
    \prob{g \mid O} = \frac{\prob{O \mid g}\prob{g}}{\sum_{g_{i} \in \goals}\prob{O \mid g_{i}}}
    \label{eq:ramirez}
\end{equation}
We develop three functions to serve as our distance metrics to measure the differnece between an observation and policy $\policy_{p}$. Since observation can be precieved in two main formats: state only list, and state and action tuples list, in this work we suggest metrics for both observations formats.

\subsection{Observation distance functions}
\label{sec:observation_distance_functions}
\noindent\textbf{Max utility for State-only}
$O^s$ applies to state-only observations and is defined by a sequence of states $<s_1, \ldots, s_k>$.  
In the AC algorithm family, the critic network estimates $V(S)$, the expected return from $s$ following a particular policy $\policy$.
$V(s)$ expresses how good it is to be in $s$, and the higher $V(s)$ the more the network thinks $s$ promise. 
We take advantage of the ability of AC to estimate $V(S)$, and present a state-only distance based on this estimation.
\begin{equation}
\label{eq:state_only_distance}
    D_s(O^s)=D_s(<s_1, \ldots, s_k>) = \frac{1}{\sum_{s_{i} \in O^s}V(s_i)}
\end{equation}

In this work, as explained in Equation \ben{Maybe delete this paragraph}
\ref{eq:softmin}, the goal that minimizes the distance between the policy and the given observation is the goal the framework suggests the agent aims.
Having said that, as the observation explains the goal better, we aim to minimize the result of distance mathematical computation.

Since in this work we use a non positive simple reward function explained in Eq \ref{eq:aggregated_reward}, as the $V(s_k \in O^s)$ becomes smaller (meaning that the state is less attractive), the state-only distance (Eq \ref{eq:state_only_distance}) becomes larger and vice versa.
If an observation contains more promising goals, their ${\sum_{s_k \in O^s}V(s_k)}$ will be larger causing $D_s(O^s)$ to become lower (sum of non positives) and eventually be assessed as a more promising goal candidate. 
% explain the formula and relation to Eq 2

\begin{figure}[t]
     \centering 
     \includegraphics[width=0.35\textwidth, height=0.15\textheight]{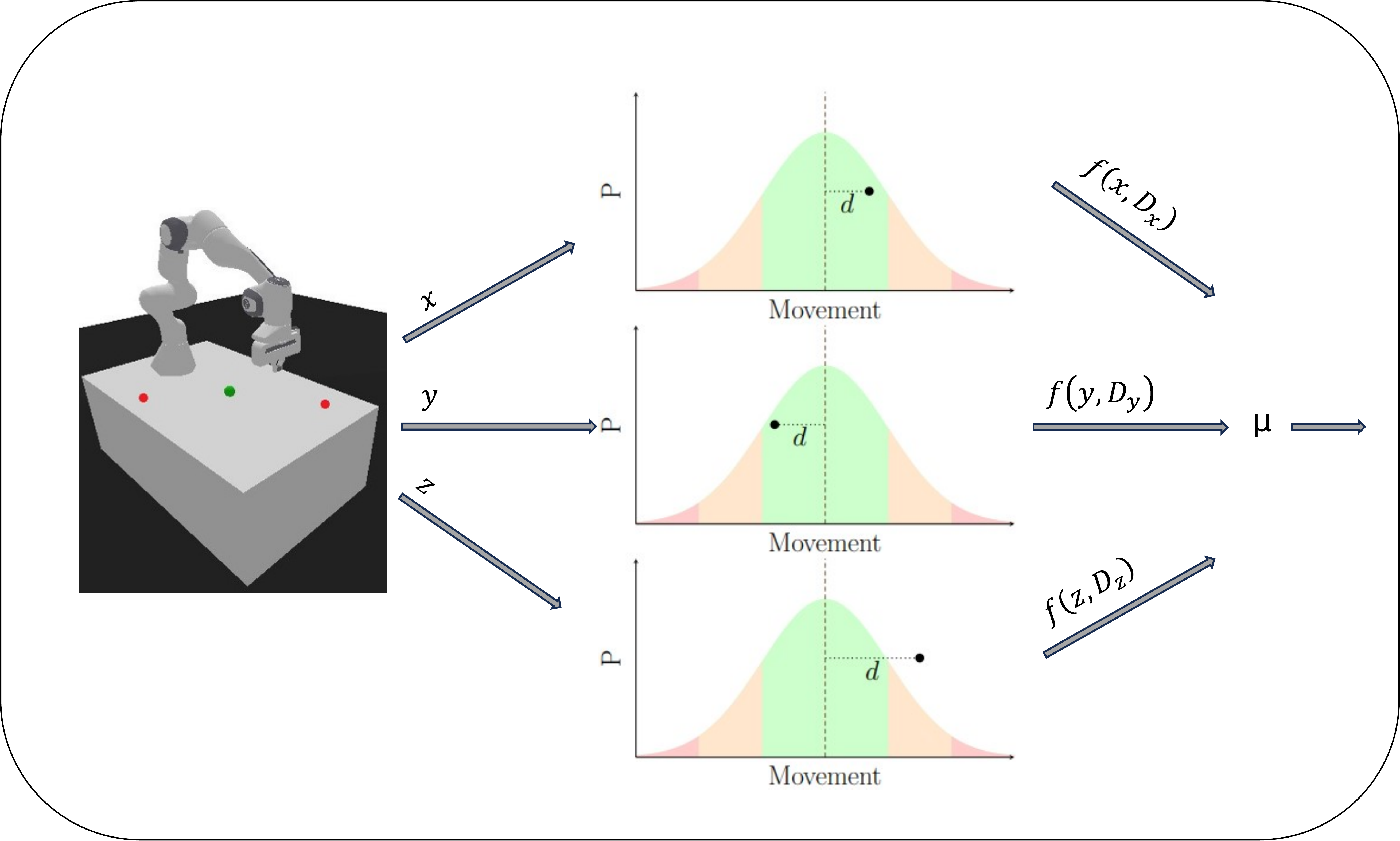}
    \caption{
    Z-score-based computation process: (i) input extraction; (ii) separately, calculate the Z-score of the agent's movement and the observation.
    (iii), return the average of calculated Z-scores.}
    \label{fig:z-score}
\end{figure}

\noindent\textbf{Wasserstein distance}
% \reuth{same here. Also, Ben, please add a citation to the relevant paper here (maybe also to Ishan's paper?)}
% \frm[inline]{Ben, like the above, use the macros $\wasserscore$ and $\wasserdistance$ to recast the formula below.}
computes the amount of work %needed %to be invested 
to convert one distribution to another \cite{vallender1974calculation}.
%Recall that the observations are likely to be sampled from a distribution. %(depending on the \textit{observers'} and \textit{actor's} configurations).
% For each state, the movement is sampled from multiple distributions (one for each axis) we calculate multiple one-dimensional Wasserstein distances (one for each axis) and report the mean. 
\begin{equation}
    \wasserscore(P, Q) = {\frac {1}{k}}\sum _{i=1}^{k}\|X_i-Y_i\|_{L_1}\label{eq:wasserstein}
\end{equation}
Eq.~\ref{eq:wasserstein} encodes the Wasserstein distance measure for one-dimensional distributions, with 1 statistical moment,
% Here \textit{p} is the number of moments. 
for two empirical distributions \textit{P} and \textit{Q} with samples $X_1, \ldots, X_k$ and $Y_1, \ldots, Y_k$ respectively.
% TODO:: BEN:: Add explanation for all variables
% we are interested in the effort to transform one dimensional distribution into another.
% \begin{flalign}
%\wasserscore(P,Q)=\left({\frac {1}{n}}\sum _{i=1}^{n}\|X_{(i)}-Y_{(i)}\|^{p}\right)^{1/p}
%\label{eq:wasserstein}
%\end{flalign}
%
We adapt this equation to compute the spot Wasserstein distance between an observation $o = s,a$ and a goal-dependent policy $\pi$, from which we sample actions $\tilde{a}$ (denoted by $\tilde \policy(s)$): 
%
%in Eq.~\ref{eq:wasserstein_spot} 
\begin{equation}
    \wasserscore(s,a,\policy)  = \| a - (\tilde{a} \sim \policy(s)) \|_{L_1}\label{eq:wasserstein_spot} 
\end{equation}
As this score looks at a single observation at a time, we omit the sum over $n$ samples.
% For each observation and agent, 
Finally, we compute the Wasserstein distance between the observation sequence and the goal-dependent policy. %in Eq.~\ref{eq:wasserstein_complete}.
% 
% Wasserstein distance is a distance function between two distributions. We compute the difference between each axis distribution's mean and the observation's action. 
\begin{align}
\wasserdistance(O, \policy) & = \mean(\{w_{i} \mid w_{i} = \wasserscore(s_i, a_i,\policy) \forall s_{i},a_{i} \in O \}) \label{eq:wasserstein_complete}
\end{align}
%
% \subsubsection{Z-score function}
\noindent\textbf{Z-score function} is a statistical measure that computes, for a value $x$ sampled from a population, the number of standard deviations $\stdev$ from the statistical means $\mean$ so that $\zscore(x) = (x - \mean)/\stdev$. As seen in Figure \ref{fig:z-score}, this computation is exactly what the inference process is looking for: how likely a sample (an observed action) is under some distribution (the goal-based policy).
Recall that policies, by default, learned via the AC methods encodes a Gaussian distribution.
% \ben{softmax?}{softmax} function of the activations from the function approximator learned during training. 
% This softmax function encodes a probability distribution from which 
Out of this Gaussian distribution, we can extract descriptive statistics, regardless of whether the action space is continuous or discrete, and leverage it to compute z-score. 
Specifically, Eq.~\ref{eq:spot-zscore} encodes the spot z-score of a single observation of state $s$, and action $a$ against the action expected of policy $\policy$ for the same state. 
\begin{equation}
    \zscore(s,a,\policy) = 
    \lvert
    \frac{a - \mean(\policy(s))}{\stdev(\policy(s))}
    \rvert
    \label{eq:spot-zscore}\\
\end{equation}
Here $\mean$, and $\stdev$ refer, respectively, to the mean and standard deviation of the numeric representation of the actions applicable in state $s$. 
We compute the distance for an entire observation sequence by taking the mean of the spot Z-scores:
\begin{equation}
    \zdistance(O, \policy) = \mean(\{z_{i} \mid z_{i} = \zscore(s_{i},a_{i},\policy) \forall s_{i},a_{i} \in O \})\label{sec:eq-zscore}
\end{equation}
% \ben{2024-01-06: Added Figure-3. Tell me if you think it would be better to have also one figure for Wasserstein or one is enough}
%
\section{Empirical Setup and Testbed}
\label{sec:results}
% \reuth{Add an algorithm showing GRDRL at work. Explain the architecture of the NN, and that you have |G| different NNs. Explain how it learns. Explain how the NN outputs are translated to compare against an observation sequence $O$. Only then show the study and the results of the study.}

% \subsection{Evaluation Domains}
\label{sec:domains}
%% FRM - I commented the subsection name out
% !TEX root = ../main.tex
We evaluate DRACO in two OpenAI-Gym~\cite{brockman2016openai} based setups: MiniGrid \cite{MinigridMiniworld23}, and Panda-gym \cite{gallouedec2021pandagym}.
As DRACO supports both discrete and continuous setups, we compare its performance with leading state-of-the-art approaches suited for these environmental setups.
For each domain, we generate GR problems by creating configurations within a shared initial setup and obstacle placement.

% \noindent\textbf{Domains:} 
% As DRACO supports both discrete and continuous setups, we compare its performance with the state-of-the-art in these environmental setups.
% We evaluate DRACO in the Minigrid domain and compare DRACO to two leading approaches: GRAQL~\cite{amado2022goal} and R\&G ~\cite{RamirezGeffner2010}. 
% \noindent\textbf{Continuous domains.} While some goal recognition algorithms can perform recognition in continuous domains, they rely on an elaborate domain representation that needs to be manually crafted \cite{vered2017online} or generated from labeled data \cite{granada2020object}. 
% Both approaches are expensive and constitute an unfair comparison for DRACO, which has none of these prerequisites. 
% Instead, to test DRACO's GR boundaries, we compare DRACO to GRAQL \cite{amado2022goal}.

\noindent \textbf{Baselines.}
(1) Plan Recognition as Planning:
Ramirez and Geffner (R\&G) is the standard baseline for symbolic GR and leverage planner executions for GR \cite{ramirez2011goal}. 
Since it relies on a PDDL domain representation that is either manually crafted \cite{vered2017online} or generated from labelled data \cite{granada2020object},
it cannot be trivially used in continuous domains, so we use it for evaluating MiniGrid.
(2) GRAQL \cite{amado2022goal} learns a Q-table per goal and uses them to compute the likelihood of each goal hypothesis. 
GRAQL's tabular nature limits it to discrete domains, so we discretize domains to compare it with DRACO in continuous setups.
\noindent\textbf{Algorithms.}  
% \ben{DRACO is a framework in which the learning algorithm might change. While we did run experiments with PPO and SAC.
% PPO performed slightly better results than SAC.}
In this evaluation, DRACO learns policies via Proximal Policy Optimization (PPO) \cite{schulman2017proximal}, from Stable Baselines \cite{raffin2021stable} and infers using of Z-score and Wasserstein distance function metrics proposed in section \ref{sec:observation_distance_functions}.
GRAQL and DRACO require offline domain's learning, whereas R\&G doesn't require training, but need to execute a planner in inference time.
This implementation exemplifies how DRACO performs with an off-the-shelf DRL algorithm. 
% Fine-tuning these networks can potentially improve the performance of DRACO, yet in this work, we show that the off-the-shelf version works sufficiently well. Moreover, we tested DRACO with Soft-Actor Critic~\cite{haarnoja2018soft}, but DRACO consistently performed better using PPO: PPO converges to a less optimal policy than with SAC, yet it still outperforms SAC, which found a policy closer to optimal. 
% This phenomenon highlights that learning a policy with a better return does not necessarily lead to a better recognition outcome; sometimes exploration is better for GR, and thus, PPO was the selected algorithm for this work's evaluation. 
% Training for DRACO's actor-critic policies use the Stable Baseline's unmodified implementation of PPO \cite{schulman2017proximal}.
% SAC consistently showed similar or slightly worse performance than PPO, so we do not report these results here. 
% The experiments use our versions of Z-score and Wasserstein distance function metrics. 
%\ben{2024-01-04: Edited this section - GRAQL uses KL-Divergence in Minigrid but mean of actions distances in continuous settings}
% For the comparison, GRAQL and DRACO require time to learn the domain, whereas R\&G requires no training, but need to execute a planner in inference time.
% We executed an unmodified version of R\&G which executed planner in inference time.
%, its distance function depends on the characteristics of the environments (the one that maximizes the GR performance). 
GRAQL uses KL-divergence for inference in MiniGrid, but its performance in discretized-continuous environments is often poor because of the high number of actions, which makes the probability that the observation and the policy converge on the same action infeasible. 
% so the probability that GRAQL converges to the exact action seen in the observation is negligible.
To address it, in the discretized environments, we use the mean of the action distances to the observation instead.
% For GRAQL, learning takes place using Q-learning, and the inference stage  leverages KL-divergence to compare each policy with a pseudo-policy based on the observations, as reported by \citet{amado2022goal}. 
% Table~\ref{tab:hyperparameters} summarizes the hyperparameters using in training.

%  \paragraph{Hyper-parameters}
% \noindent\textbf{Hyper-parameters.} 
% In Minigrid \ref{sec:minigrid}, PPO's hyperparameter values are: number of episodes is 100k, discount factor ($\gamma$) is 0.99 and the learning rate ($\alpha$) is 0.001.
% The $\epsilon$ used for computing $\pi_O$ is $1e^{-6}$. 
% However, in Panda-Gym \ref{sec:panda} both SAC and PPO share the RL hyperparameters: number of episodes (25k), discount factor $\gamma$ 0.93, the learning rate ($\alpha$) is 0.0001.
% Nevertheless, these algorithms do not share all of the hyperparameters. 
% PPO's clip range is 0.3, batch size 64 while SAC's batch size is 512.

% In Minigrid \ref{sec:minigrid}, PPO's hyperparameter values are: number of episodes is 100k, discount factor $\gamma = 0.99$ and the learning rate $\alpha = 0.001$.
% The $\epsilon$ used for computing $\pi_O$ is $1e^{-6}$. 
% Hyperparameters for Panda-Gym \ref{sec:panda} are: number of episodes (25k), discount factor $\gamma=0.93$, the learning rate $\alpha = 0.0001$, and clip range is $0.3$.
% \ben{2023-01-04: Added Hyper-parameters section instead of the table} \reuth{Sorry, I probably missed that. Does this change save space?}
\noindent\textbf{Hyperparameters.}
While GRAQL and DRACO use different training algorithms, they share various hyperparameters, such as number of episodes, learning rate, and discount factor. 
Throughout the executions, optimizing the approach's performance was the leading motivation. 
We used the hyperparameter values that maximized each approach. % 
\ben{Lets move hyperparameter details to the supplent files}In MiniGrid, we compare GRAQL and DRACO after $100K$ episodes, discount factor ($\gamma$) $0.99$, and learning rate ($\alpha$) $0.001$.
In Panda-Gym, we tracked the learning's performance under multiple episode configurations until convergence. 
The optimal discretization factor for GRAQL is $0.03$ cm. Similarly, in Panda-Gym, GRAQL's learning rate ($\gamma$) was $0.01$ while DRACO's was $0.0006$. 
For R\&G, we created a PDDL-based domain theory using PDDL-generator \cite{seipp-et-al-zenodo2022}.

% \begin{table}[b]
%     \centering
%     \small
%     \footnotesize
%     \setlength{\tabcolsep}{2pt}
%     \begin{tabular}{|l|l|l|l|l|l|l|l|l|} 
%     \cline{2-9}
%     \multicolumn{1}{c|}{} & \multicolumn{4}{c|}{GRAQL}                     & \multicolumn{4}{c|}{DRACO}                      \\ 
%     \cline{2-9}
%     \multicolumn{1}{l|}{} & $\alpha$ & $\epsilon$ & $\gamma$ & Epis. & $\alpha$ & $\epsilon$ & $\gamma$ & Epis.  \\ 
%     \hline
%     Minigrid              & $1e^{-3}$      & $1.5e^{-7}$     & $0.93$       & $50k$   & $1e^{-3}$      & $1e^{-6}$     & $0.93$       & $25k$    \\ 
%     \hline
%     Panda-Gym             & $1e^{-4}$     & $0.3$          & $0.99$       & $100k$  & $1e^{-4}$     & $0.3$          & $0.99$       & $100k$   \\
%     \hline
%     \end{tabular}
%     \normalsize
%     \caption{Hyperparameters for Experimentation}
%     \label{tab:hyperparameters}
% \end{table}

% \paragraph{Metrics.} 
\noindent\textbf{Metrics.} 
% The algorithms are evaluated on standard prediction-quality metrics often used in machine learning research. 
% When used in the context of goal recognition, to ``correctly classify'' the true goal means it was given the highest probability. 
% As such, the meaning of these metrics for GR is as follows: Precision measures how often was the predicted most likely goal the correct goal; Accuracy measures how often we correctly classified the correct and incorrect goals, and recall measures how many times the correct goal was identified as the most likely goal.
We leverage standard prediction metrics from machine learning, to the context of goal recognition, to ``correctly classify'' the true goal means it has the highest probability. 
As such, the meaning of these metrics for GR is as follows: 
% Precision measures how often a method predicts the correct goal as most likely;
Accuracy and precision measure how often a method correctly classifies the correct and incorrect goals, respectively, and recall measures how often a method identifies the correct goal as the most likely.
Finally, we evaluate the framework's confidence in its prediction. 
The confidence metric is the difference between the probabilities of the two most probable goals divided by the most probable goal's probability. 
Thus, if two (or more) most likely goals are \emph{equiprobable}, confidence is \emph{zero}. 
% This means that if a method outputs similar probabilities for two goals, its confidence tends to zero, whereas if it outputs a substantially higher probability to a single goal, its confidence tends to one. 
% This confidence metric calculates the relative distance between the predicted goal and the second-most probable goal candidate divided by distance. 
% \leo[inline]{In GRAQL we follow the machine learning notation. Thus, it would be something like this:
% - Precision: How often was the predicted most likely goal the correct goal? 
% - Accuracy: How often did we correctly classify the correct and incorrect goals? (This is useful when dealing with spread)
% - Recall: How many times were the correct goals identified as the most likely goal?
% }
% \begin{itemize}
% \item Precision: How often was the predicted most likely goal the correct goal? 
% \item Accuracy: How often did we correctly classify the correct and incorrect goals? 
% \item Recall: How many times were the correct goals identified as the most likely goal?
% \item F-score: the standard computation with respect to the other metrics.
% % \begin{equation*}
% %     \frac{precision \times recall}{precision + recall}
% % \end{equation*}
% \end{itemize}

% To enable this tabular approach to work in Panda-gym, we provide a discretization of that environment as will be discussed in Section \ref{sec:panda}. %% FRM - So let's go into details in that section

\begin{figure}[tb]
     \centering
     \begin{subfigure}[b]{0.14\textwidth}
         \centering
         \includegraphics[width=\textwidth]{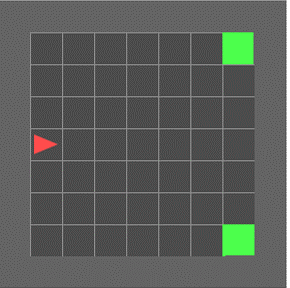}
         \caption{2-Goal.}
         \label{fig:2goal}
     \end{subfigure}
     % \hfill
     \begin{subfigure}[b]{0.14\textwidth}
         \centering
         \includegraphics[width=\textwidth]{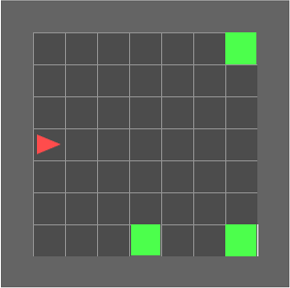}
         \caption{3-Goal.}
         \label{fig:3goal}
     \end{subfigure}
     \begin{subfigure}[b]{0.14\textwidth}
         \centering
         \includegraphics[width=\textwidth]{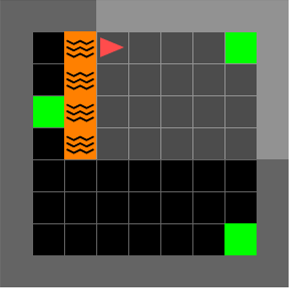}
         \caption{Lava.}
         \label{fig:lava}
     \end{subfigure}
     % \hfill
          \begin{subfigure}[b]{0.14\textwidth}
         \centering
         \includegraphics[width=\textwidth]{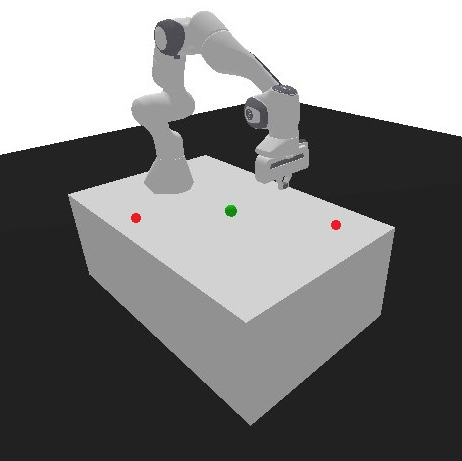}
         \caption{Panda-gym.}
         \label{fig:panda_arm_env}
     \end{subfigure}
        \caption{Example for domain setups used in evaluation. 
        % In MiniGrid, the agent, goal candidates and lava cells are marked with a red triangle, green cells, and orange cells accordingly. In Panda-gym, the robotic-arm can reach one out of three goal candidates denoted by red or green sphere.
        }
    \label{fig:minigridpics}
\end{figure}

\noindent\textbf{Domain 1: MiniGrid.} 
% \subsubsection{Domain 1: MiniGrid}
\label{sec:minigrid}
% MiniGrid is a library containing a collection of discrete grid-world environments for goal-oriented tasks. 
%, and configurable obstacles, termination states, agent positions, and settings. 
% Obstacles in MiniGrid can vary from one to another: a wall blocks the agent's movement, while a door blocks until the agent picks up its keys.
The motivation to use MiniGrid is the ability to be represented in multiple formats and to model complex GR problems.
We created three environment formats, one for each algorithm: PDDLs for R\&G (complete environment model), symbolic state representation for GRAQL (x and y coordinates, angle) and a visual representation for DRACO (image), making it possible to compare the different approaches in the same GR problem.
Note that each representation varies in the information each algorithm perceives. 
% The PDDL input of R\&G is a complete model of the environment, the symbolic representation provides more structured data to the recognizer, as it parses the state into semantically meaningful components (x and y coordinates, angle), and DRACO uses raw-state representation.
% supports both a symbolic state representation, and a visual representation. 
% The visual representation is easier for NNs, as the ones used in the DRACO framework, to process. \frm{Not sure what you mean by ``to process here''}
% The symbolic state representation enables us to compare this work with existing GR algorithms on the exact same scenarios, without the need to artificially translate the visual observations or use an external activity-recognition module. 
% \begin{figure}[tb]
%     \centering
% \includegraphics[width=0.4\textwidth]{Figures/panda-3-goals-green-middle.jpg}
%     \caption{Panda-gym environment with 3 goal locations.}
%     \label{fig:panda_arm_env}
% \end{figure}
MiniGrid problem consists of an agent and objective which are the red triangle and the bright green cells in Figure \ref{fig:minigridpics} respectively. The agent's task is to reach the objective by taking actions: \textit{turn right, turn left, move forward, do nothing}. 
We evaluate the three approaches on the setups depicted in Figures \ref{fig:2goal}, \ref{fig:3goal}, \ref{fig:lava}: 2-Goal and 3-Goal are obstacle-free environments, and Lava has 3 goal candidates and lava cells. For each setup, we generated 10 different GR settings.
\noindent\textbf{Domain 2: Panda-gym} 
% \subsubsection{Domain 2: Panda-gym}
\label{sec:panda}
% Panda-gym is a library containing multiple continuous environments for goal-oriented tasks. 
Panda-gym is a drastically different domain than MiniGrid: it has continuous state and action spaces, and it realistically presents a Franka Emika Panda robot arm \cite{gallouedec2021pandagym} using the PyBullet physics engine.  
This domain allows us to test DRACO's boundaries and performance on a continuous state space that is not an image.
% We use Panda-gym because its action and observation spaces are continuous, and it realistically presents a Franka Emika Panda robot arm \cite{gallouedec2021pandagym} using the PyBullet physics engine. 
This environment consists of a single arm with the objective of reaching a colored sphere (Figure~\ref{fig:panda_arm_env}).
% Similarly to our modifications in MiniGrid from Section~\ref{sec:minigrid}, we modified the environment to simulate GR problems by accumulating several goal configurations into a single task with numerous goal candidates. %\frm{This is a bit repetitious, we need more detail here. Perhaps illusrate multiple scenarios, like we did with minigrid.}
Since this environment is fully continuous and does not inherently support discrete state spaces, we provide a discretization of the environment in order to run the GRAQL baseline on it. 
Specifically, we tested discretization granularities of $0.01-0.05$ cm and eventually chose $0.03$ cm as this value yields the best performance for our GRAQL baseline. 
% We tune the granularity of the discretization and use the values for which GRAQL provides the best results. 
% The values we tested are $0.01, 0.02, 0.03, 0.05$ and we eventually chose $0.03$ because it achieved the best learning results. \reuth{The values we tested we .... We eventually chose...} 
% The evaluation in this environment aims to showcase DRACO's boundaries and performance on a continuous state space that is not an image, rather than compare it to the current state-of-the-art, which is unable to process this unadjusted domain altogether.
% In this comparison between the two approaches we chose the optimal discretization factor which 
To stress out approach, we ran both DRACO and GRAQL on three different GR setups, with 2,3, and 4 goal spheres (Figure \ref{fig:panda_arm_env} illustrates Panda-Gym 3-goals problem). For each setup, we generated 10 different GR settings.

% \noindent\textbf{Domains ambiguity levels} 
% The setups are designed to have varying levels of ambiguity. Notably, in 3 and 4 goals, the movements towards the goal in some partial observability conditions (mainly 10\% and 30\%) can even be impossible to disambiguate. 
% For each instance we generated 10 different GR settings.

% We measure the GR problems ambiguity level using an equivalent computation to the \textit{WCD} as in Minigrid's ambiguity evaluation, but as \textit{WCD} accounts for discrete states and Panda-Gym is a continuous environment, we considered all states in a perimeter of $0.01$ cm around a state to be the same. In the \textit{2-Goals}, \textit{3-Goals} and \textit{4-Goals} Panda-Gym setups, the WCD was 1, 4, and 7, which is $7\%$, $57\%$, and $85\%$ of the optimal path respectively.

\noindent\textbf{Domain ambiguity} \label{par:domain_ambiguity} We assess the 
level of domain ambiguity using the \textit{ worst-case distinctiveness (WCD)} metric \cite{keren2014goal}. WCD represents the maximum steps an optimal agent can take before an \textit{observer} can differentiate the correct goal.
Since WCD accounts only for discrete states, in Panda-gym, we consider all states in a perimeter of $0.01$ cm around a state to be the same.
Table \ref{tab:wcdEnv} summarizes the WCD on evaluated domains, and shows that across all domains, we specifically chose ambiguous setups in which it would be challenging to perform GR.
Notably, in Panda-Gym 3 and 4 goals and in MiniGrid 2 and 3 goals problems, the movements towards the goal under some partial observability conditions (mainly 10\% and 30\%) can even be impossible to disambiguate.

\begin{table}[]
\centering
\footnotesize
\tiny
\begin{tabular}{|c|c|c|c|c|c|c|c|}
\hline
Domain & Problem & WCD & WCD / |$p_g$| & Domain & Problem & WCD ($\delta = 0.01cm$) & WCD ($\delta = 0.01cm$) / |$p_g$| \\ \hline
\multirow{3}{*}{MiniGrid} & Lava & 5 & 5/11 $\approx$ 45\% & 
\multirow{3}{*}{Panda-Gym} & 2-Goals & 1 & 1/14 $\approx$ 7\% \\ 
\cline{2-4} \cline{6-8} & 2-Goals & 6 & 6/9 $\approx$ 66\% & & 3-Goals & 4 & 4/7 $\approx$ 57\% \\ \cline{2-4} \cline{6-8} & 3-Goals & 6   & 6/9 $\approx$ 66\%     & & 4-Goals & 7 & 7/8 $\approx$ 87\% \\ 
\hline
\end{tabular}

\caption{WCD metric for evaluated domains and problems. $p_g$ is an optimal plan for a goal g. $\delta$ is size of the perimeter we consider states to be the same (for continuous environments). }
\label{tab:wcdEnv}

\end{table}

\begin{table*}[tb]
\footnotesize
\tiny
\setlength{\tabcolsep}{2pt}
    \centering
    \begin{tabular}{|c|c|c|c|c|c|c|c|c|c|c|c|c|c|c|c|c|c|}  
\hline  
& &
 \multicolumn{5}{|c|}{DRACO using KL-Divergence} & \multicolumn{5}{|c|}{GRAQL using KL-Divergence} &
 \multicolumn{5}{|c|}{Ramirez and Geffner} \\
\hline  
OBS & Problem & Accuracy & Precision & Recall & F-Score & Conf. & Accuracy & Precision & Recall & F-Score & Conf.& Accuracy & Precision & Recall & F-Score & Conf.\\ 
\hline  
10\% & Lava &
\multicolumn{1}{c|}{0.73 \textpm 0.32} &
\multicolumn{1}{c|}{0.6 \textpm 0.52} &
\multicolumn{1}{c|}{0.6 \textpm\ 0.52} &
\multicolumn{1}{c|}{0.6 \textpm 0.52} &
\multicolumn{1}{c|}{\textbf{81\%}} &
\multicolumn{1}{c|}{0.53 \textpm\ 0.32} &
\multicolumn{1}{c|}{0.41 \textpm\ 0.36} &
\multicolumn{1}{c|}{0.9 \textpm\ 0.32} &
\multicolumn{1}{c|}{0.56 \textpm\ 0.32} &
\multicolumn{1}{c|}{5\%} &
\multicolumn{1}{c|}{$\textbf{1.0}\boldsymbol{\pm} \textbf{0.0}$} &
\multicolumn{1}{c|}{$\textbf{0.77} \boldsymbol{\pm} \textbf{0.46}$} &
\multicolumn{1}{c|}{$\textbf{1.0}\boldsymbol{\pm} \textbf{0.0}$} &
\multicolumn{1}{c|}{$\textbf{0.87} \boldsymbol{\pm} \textbf{0.46}$}  &
\multicolumn{1}{c|}{70\%}
\\
 & 2-Goals &
 \multicolumn{1}{c|}{$\textbf{1.0}\boldsymbol{\pm} \textbf{0.0}$} &
 \multicolumn{1}{c|}{$\textbf{1.0}\boldsymbol{\pm} \textbf{0.0}$} &
 \multicolumn{1}{c|}{$\textbf{1.0}\boldsymbol{\pm} \textbf{0.0}$} &
 \multicolumn{1}{c|}{$\textbf{1.0}\boldsymbol{\pm} \textbf{0.0}$} &
\multicolumn{1}{c|}{\textbf{77\%}} & 
\multicolumn{1}{c|}{0.70 \textpm\ 0.35} &
\multicolumn{1}{c|}{0.64 \textpm\ 0.35} &
\multicolumn{1}{c|}{0.9 \textpm\ 0.31} &
\multicolumn{1}{c|}{0.75 \textpm\ 0.32} &
\multicolumn{1}{c|}{13\%} &
\multicolumn{1}{c|}{$\textbf{1.0}\boldsymbol{\pm} \textbf{0.0}$} &
\multicolumn{1}{c|}{0.5 \textpm\ 0.0} &
\multicolumn{1}{c|}{$\textbf{1.0}\boldsymbol{\pm} \textbf{0.0}$} &
\multicolumn{1}{c|}{0.67 \textpm\ 0.00} &
\multicolumn{1}{c|}{0\%}
\\  
 & 3-Goals &
 \multicolumn{1}{c|}{$\textbf{1.0}\boldsymbol{\pm} \textbf{0.0}$} &
 \multicolumn{1}{c|}{$\textbf{1.0}\boldsymbol{\pm} \textbf{0.0}$} &
 \multicolumn{1}{c|}{$\textbf{1.0}\boldsymbol{\pm} \textbf{0.0}$} &
 \multicolumn{1}{c|}{$\textbf{1.0}\boldsymbol{\pm} \textbf{0.0}$} &
\multicolumn{1}{c|}{\textbf{78\%}} &
\multicolumn{1}{c|}{0.66 \textpm\ 0.35} &
\multicolumn{1}{c|}{0.56 \textpm\ 0.40} &
\multicolumn{1}{c|}{0.9 \textpm\ 0.32} &
\multicolumn{1}{c|}{0.64 \textpm\ 0.35} &
\multicolumn{1}{c|}{6\%} &
\multicolumn{1}{c|}{$\textbf{1.0}\boldsymbol{\pm} \textbf{0.0}$} &
\multicolumn{1}{c|}{0.36 \textpm\ 0.0} &
\multicolumn{1}{c|}{$\textbf{1.0}\boldsymbol{\pm} \textbf{0.0}$} &
\multicolumn{1}{c|}{0.52 \textpm\ 0.00} &
\multicolumn{1}{c|}{0\%}\\
 \hline  
30\% & Lava & 
\multicolumn{1}{c|}{0.93 \textpm 0.21} &
\multicolumn{1}{c|}{0.9 \textpm 0.31} &
\multicolumn{1}{c|}{0.9 \textpm 0.31} &
\multicolumn{1}{c|}{0.9 \textpm 0.31} &
\multicolumn{1}{c|}{\textbf{64\%}} &
\multicolumn{1}{c|}{0.87 \textpm\ 0.28} &
\multicolumn{1}{c|}{0.80 \textpm\ 0.42} &
\multicolumn{1}{c|}{0.80 \textpm\ 0.42} &
\multicolumn{1}{c|}{0.80 \textpm\ 0.42} &
\multicolumn{1}{c|}{8\%} & 
\multicolumn{1}{c|}{$\textbf{1.0}\boldsymbol{\pm} \textbf{0.0}$} &
\multicolumn{1}{c|}{$\textbf{1.0}\boldsymbol{\pm} \textbf{0.0}$} &
\multicolumn{1}{c|}{$\textbf{1.0}\boldsymbol{\pm} \textbf{0.0}$} &
\multicolumn{1}{c|}{$\textbf{1.0}\boldsymbol{\pm} \textbf{0.0}$} &
\multicolumn{1}{c|}{55\%} 
\\  
 & 2-Goals & 
 \multicolumn{1}{c|}{$\textbf{1.0}\boldsymbol{\pm} \textbf{0.0}$} &
 \multicolumn{1}{c|}{$\textbf{1.0}\boldsymbol{\pm} \textbf{0.0}$} &
 \multicolumn{1}{c|}{$\textbf{1.0}\boldsymbol{\pm} \textbf{0.0}$} &
 \multicolumn{1}{c|}{$\textbf{1.0}\boldsymbol{\pm} \textbf{0.0}$} &
\multicolumn{1}{c|}{\textbf{65\%}} &
\multicolumn{1}{c|}{$\textbf{1.0}\boldsymbol{\pm} \textbf{0.0}$} &
\multicolumn{1}{c|}{$\textbf{1.0}\boldsymbol{\pm} \textbf{0.0}$} &
\multicolumn{1}{c|}{$\textbf{1.0}\boldsymbol{\pm} \textbf{0.0}$} &
\multicolumn{1}{c|}{$\textbf{1.0}\boldsymbol{\pm} \textbf{0.0}$} &
\multicolumn{1}{c|}{12\%} & 
\multicolumn{1}{c|}{$\textbf{1.0}\boldsymbol{\pm} \textbf{0.0}$} &
\multicolumn{1}{c|}{0.71 \textpm\ 0.49} &
\multicolumn{1}{c|}{$\textbf{1.0}\boldsymbol{\pm} \textbf{0.0}$} &
\multicolumn{1}{c|}{0.83 \textpm\ 0.48} &
\multicolumn{1}{c|}{30\%} 
\\  
 & 3-Goals &
 \multicolumn{1}{c|}{$\textbf{1.0}\boldsymbol{\pm} \textbf{0.0}$} &
 \multicolumn{1}{c|}{$\textbf{1.0}\boldsymbol{\pm} \textbf{0.0}$} &
 \multicolumn{1}{c|}{$\textbf{1.0}\boldsymbol{\pm} \textbf{0.0}$} &
 \multicolumn{1}{c|}{$\textbf{1.0}\boldsymbol{\pm} \textbf{0.0}$}&
\multicolumn{1}{c|}{\textbf{38\%}} &
\multicolumn{1}{c|}{$\textbf{1.0}\boldsymbol{\pm} \textbf{0.0}$} &
 \multicolumn{1}{c|}{$\textbf{1.0}\boldsymbol{\pm} \textbf{0.0}$} &
 \multicolumn{1}{c|}{$\textbf{1.0}\boldsymbol{\pm} \textbf{0.0}$} &
 \multicolumn{1}{c|}{$\textbf{1.0}\boldsymbol{\pm} \textbf{0.0}$} &
\multicolumn{1}{c|}{11\%} &
\multicolumn{1}{c|}{$\textbf{1.0}\boldsymbol{\pm} \textbf{0.0}$} &
 \multicolumn{1}{c|}{0.62 \textpm\ 0.49} &
 \multicolumn{1}{c|}{$\textbf{1.0}\boldsymbol{\pm} \textbf{0.0}$} &
 \multicolumn{1}{c|}{0.77 \textpm\ 0.48} &
\multicolumn{1}{c|}{30\%}
\\
\hline
\end{tabular}
    \caption{
        % Impact of partial observability, KL-Divergence is used to measure the distance between the optimal paths and the given observation in both approaches. The results are the average (± standard deviation) of 10 different executions.
        % Performance in the Minigrid domain under 10\% and 30\% partial observability.
        % Impact of partial observability on the performance of the three approaches in the MiniGrid domain. 
        The table reports the average performance (± standard deviation) of the algorithms over 30 different GR settings under 10\% and 30\% partial observability. \textit{Conf.} stands for Confidence}
    \label{tab:minigrid}
\end{table*}

\section{Results}
% In discrete settings, we compare DRACO to both Goal
% To assess the performance of DRACO in continuous settings we first compare it to GRAQL and then discuss the different metrics proposed in section \ref{sec:infer}. 
Our tests use a different algorithm from our learning methods to generate observations, namely Asynchronous Actor-Critic (A2C) \cite{mnih2016asynchronous}.
For each combination of algorithm and instance, the observation sequence $O$ %provided to the algorithms 
varies its degree of observability: $10\%$, $30\%$, $50\%$, $70\%$, and full observability. 
We generate an observation sequence with $x\%$ observability by taking a full sequence and removing each observation along the trajectory with $x\%$ probability. 
Similarly, we generate noisy observations by appending suboptimal actions to $x\%$ of the states.\footnote{Examples of noisy observations in the supplement.} 
For each execution in every environment, we use the same trajectories for all algorithms.
We show the metrics averaged over $10$ settings with varying degrees of observability ($13$ configurations per execution), totalling $540$ GR problem instances.
% In total, $540$ GR problem instances were executed on Minigrid and Panda-Gym. 
% and their respectively summarized results appear in Table~\ref{tab:minigrid} and Figure~\ref{fig:panda-comp}.

\noindent\textbf{Performance} 
\ben{Maybe split between minigrid and Panda results}
% \subsubsection{Performance} 
% , both in MiniGrid and Panda-gym. 
% For each execution we used the same trajectories in the two frameworks, removing steps randomly to achieve a specific observability ratio. 
% \input{sections/minigrid-table2}
Table~\ref{tab:minigrid} shows results for MiniGrid. 
All approaches perform overwhelmingly well, with perfect scores on all measures when the observability is $0.5$ or higher. % or equal to 0.5, both frameworks achieved perfect scores.
However, this result misses part of the story: %, as both of the algorithms calculated the difference between the observation and the policy for each goal GRACO's distance measurement was more accurate.
while GRAQL's and R\&G's true goal confidence score was close to the other goals' score ($9\%$ and $13\%$ difference on average respectively), DRACO had very different scores between the true goal and the other candidates ($71\%$ on average), making DRACO more confident in its predictions. 
% making its confidence in the true goal higher. 
Moreover, DRACO's results dominate GRAQL's and R\&G results in the more challenging, low observability ($10\%$, $30\%$), problem variants. 
It is a reasonable outcome because DL can make significant inroads to recognizing patterns and variance instead of valuing the current state. 
Moreover, our use of NN-based policies makes DRACO more forgiving to missing observations, as the likelihood of these missing actions is $\epsilon$ rather than zero, and this difference is accumulated with the number of missing observations.
% Lastly, when evaluating the algorithm's performance on noisy observation sequences, the results show that both GRAQL and DRACO were able to correctly identify the true goal of the agent. 
%
% However, even 
%
% However, This result is misleading, as DRACO ranked the true goal with a much higher probability than the other goals, while in GRAQL the difference between the true goal and the others was smaller.
%
%
Lastly, as this table shows, DRACO and GRAQL both outperform the R\&G approach in most of these discrete domains. 
Importantly, DRACO dominates both approaches in the confidence metric, inferring higher probabilities for the correct goal, whereas GRAQL and R\&G often disambiguate poorly. 
% which leverages planning.
% So, while a direct comparison between DRACO and PDDL-based approaches is not trivial, a comparison to an intermediate solution (GRAQL) shows that DRACO outperforms it. 
%by matching up to the performance of GRAQL, DRACO can be considered to transitively outperform these methods as well.
% 
Note that the accuracy metric in this paper differs from \citet{RamirezGeffner2010}, which refers to the ratio of problems where the recognizer ranks the true goal (i.e., True Positives) with the highest likelihood, and the number of problems tested. 
This measure differs from standard ML accuracy, which also factors in the successful ranking of the correct goal (i.e., True Positives + True Negatives). 
In our experiments, we report accuracy using the latter definition.

Figure~\ref{fig:panda-comp} reports the average F-Score for all Panda-Gym problems running GRAQL in a discretized environment version and DRACO in the continuous environment representation, both with the Z-score and the Wasserstein distance metrics. Figure~\ref{fig:panda-f-score-episodes} shows the mean performance with full observability, and Figures~\ref{fig:panda-f-scoreobservability}-~\ref{fig:noisy_results} show the mean performance under partial and noisy observations.
%
%using the proposed evaluation metrics for inference.
First, DRACO's performance is similar and often superior to GRAQL, often reaching the best results regardless of the distance metric used. GRAQL struggles in generalizing continuous and relatively large environments. This gap happens due to one of the main differences between DL and tabular methods --- DL approximates states similar to visited ones while Q-Learning does not, affecting the number of episodes GRAQL requires to learn how to solve a problem.
%
% Indeed, some exploration to get policy estimations for a larger portion of the state space is desirable. %\reuth{Move the contrastive discussion here?}
% DRACO with the two proposed distance metrics, Wasserstein and Z-score, %, with the Wasserstein distance metric being slightly more informative for the DRACO-SAC variant. 
%
Figures~\ref{fig:panda-f-score-episodes} and~\ref{fig:panda-f-scoreobservability} compare the approaches under increasing episodes and observability percentage, respectively.
\footnote{Supplementary material includes a breakdown of the results.}
These figures emphasize the superiority of DRACO over GRAQL across the board in continuous environments. GRAQL performance improves when episodes increase but generalizing a large problem for GRAQL takes a substantially more amount of computations than it takes for DRACO. However, in the partial observation scenario (Figure \ref{fig:panda-f-scoreobservability}), as the observability percentage increases DRACO's performance increases as well while GRAQL's remains about the same, thus emphasizing its inability to learn the problem (even for the 100\% observability problem).
Figure~\ref{fig:noisy_results} illustrates performance under noisy (but complete) observations. DRACO outperforms GRAQL with low variance, showing its substantial resilience to noise. DRACO was barely impacted by the 10\% and 20\% noise ratio, preserving its performance from the full-observable non-noisy observation.
Indeed, the gap between DRACO and GRAQL increases as the noise ratio decreases. 
Such a difference in performance is again likely due to its leverage of DL to the learning stage and the richer representation state available to DRACO.
% In our test environment, overall, DRACO and GRAQL achieve similar results in the discrete environment  high observability. However, in both low observability discrete environments and off-course continuous environments DRACO achieves higher performance with lower variance.
%
Additionally, Figure \ref{fig:panda-comp} also compares the different distance metrics proposed in this paper in section \ref{sec:infer}. Figure \ref{fig:panda-f-score-episodes} shows a slight dominance of DRACO using Wasserstein distance over DRACO using Z-score when trained for $ < 125K$ but for 150K+ episodes, this trend changes.
The number of trained episodes impacts the agent's optimality, making the Z-score metric better for scenarios where the trained agent policy is more optimal.
This behavior happens because the Z-score metric takes into consideration the standard deviation of the policy distribution which takes more episodes to stabilize.
Figures \ref{fig:panda-f-scoreobservability}, \ref{fig:noisy_results} shows a dominance of Wasserstein over Z-score, for observability
levels of $< 50\%$ and with noisy observations.
Wasserstein is better under high uncertainty stemming from the observations, as the Z-score metric is more sensitive to high variability that skews it from the right goal.

\input{sections/minigrid-figures}
%
% \ben{Added the Scalability Paragraph - feel free to modify}
\noindent\textbf{Scalability.} 
% \subsubsection{Scalability} 
Scalability evaluation has three main factors: storage, training, and execution time. 
First, in terms of storage consumption, DRACO uses a constant $8 \pm 0$ MB while GRAQL $3190 \pm 700$ MB. 
This difference is primarily due to the huge storage needs of the Q-table used by GRAQL.
% (which is depends on the discretization factor hyperparameter).
% The Q-table's size depends on the hyperparameter discretization factor. As the descritization factor decrease the number of valid states increases (which impact the Q-table size significantly).
% number inscrease selection because the number of valid states and possible actions in these states increase as we decrease this factor.
% This difference derives from the usage of Q-Learning by GRAQL. Q-Learning saves a cell in its table for every unique state and action tuple, making it tremendously heavy especially for large domains like Panda-Gym. 
By contrast,  DRACO uses constant memory because it stores only the network architecture and neurons' values. Even though these neurons' values change along the training time, their memory consumption remains constant.
% However, DRACO uses constant memory because it stores only the network architecture and neurons' values. 
% In terms of the second factor of scalability, execution time, 
Second, DRACO's average training time was $13.52 \pm  4.3$ minutes while GRAQL's was $1108 \pm 463$ minutes. We trained both approaches on the same commodity Tesla K40 server.
% This difference can be related to the exhaustive use of storage by GRAQL and also related to GRAQL's implementation which relies on the CPU computation only.
This difference is due to two factors: (1) GRAQL relies on the CPU for its computation, whereas DRACO uses a GPU; and (2) GRAQL's storage-intensive operations are more expensive. Notably, R\&G does not require training time, yet its online time is significantly higher than the other algorithms, as it runs planners in inference time. Specifically, DRACO and GRAQL's online runtime takes $\approx0.12$ and $\approx0.1$ seconds on average respectively while R\&G takes $\approx2$ seconds.
Third, as discussed in section \ref{sec:infer}, GRAQL and DRACO load their policies before inference. GRAQL's Q-tables loading time is often not negligible and takes an average of $\approx 3.4$ minutes per table while DRACO's NNs loading takes seconds.
% Note that both approaches were trained on the same commodity Tesla K40 server.

To conclude, we evaluate the paper's two main contributions: the DRACO framework and the proposed distance metrics.
DRACO scaled better than the state-of-the-art GR algorithms in storage consumption, computation time, and performance.
DRACO requires no domain expert to craft an elaborate model like R\&G and instead learns via interactions.
It achieves better or equal performance results compared to GRAQL and R\&G, with both metrics performing overwhelmingly well. 
Wasserstein was slightly better in less observable spaces and under noisy observations.

% In our test environment, DRACO always recognizes the correct goal (exclusively so), and with higher probability.%, for all problems.

%To conclude, these preliminary results exemplify two out of the three different scenarios discussed in this proposal in section \ref{sec:hypotheses}. These results provide initial support to the first and second hypotheses.
%The other hypothesis is yet to be researched in the preliminary results but will further be researched in this work. 
%\leo[inline]{Removed text referring to Ben's proposal, will later include a conclusion to this section}

 \section{Related Work}\label{sec:related_work}
     % !TEX root = ../main.tex

Goal recognition research can be categorized into two main categories: symbolic and data-driven GR.  
Symbolic approaches leverage planning, search, and parsing techniques to infer which goal best explains a sequence of symbols~\cite{geib2009probabilistic,pereira2020landmark}. %% TO add if accepted: SLIM.
These approaches heavily rely on a domain expert to provide a problem and environment formulation, which makes them difficult to scale and susceptible to noise.
%Moreover, traditional GR algorithms require complex computation processes, such as multiple planner runs, to calculate the similarity of the observability to each possible goal.
% Adjusting to new domains is very costly and might be also difficult because it requires adjusting and defining the problem's planner, parser, and solution from the beginning.
% \frm{We spend a lot more space on the data-driven approaches, perhaps we should rebalance. }
% \subsubsection{Data-Driven GR}
%\frm[inline]{Consider moving this entire paragraph to another section at the end. We have plenty of space. But I will hold off the change until we have the entire paper here.}
% An alternative approach is data-driven GR. 
By contrast, data-driven GR obviates or mitigates the need for a domain expert. %% TO add if accepted: new survey
Some learning-based approaches rely on process mining techniques \cite{polyvyanyy2020goal,Ko2023plan}. 
These approaches compare an observation sequence to past traces, which makes them brittle when handling large or continuous action spaces. 
Recent work lifts the requirement of domain experts by relying on tabular RL or DL \cite{zeng2018inverse,min2014deep}.
%\frm{We may wish to fold these two approaches together (ours and Zeng's) since both rely on RL (tabular?). But the question then is, why do we not test with Zeng's approach?} \reuth{Because Zeng et al. uses IRL and requires a database of real interactions.}\frm{So it does require a domain expert. }
\citet{amado2022goal} develop a GR approach based on tabular RL techniques to learn the domain model, and 
\citet{fang2023real} extend this work to robotics simulators. 
Like many other RL algorithms, these algorithms assume that the agent's environment is an MDP.
%This work also proposes multiple semantics for the term "explains" \cite{amado2022goal}  as presented by Ramirez and Geffner \cite{ramirez2009plan} definition to GR and takes a measure-based approach instead of running planners to compare different goals. For each goal, the framework calculates the distance between the goal's Q-Table and the observation. The chosen goal is the one that minimizes this distance. In addition, they used three off-the-shelf distance measure functions (KL divergence, MaxUtil, and Divergence Point). This approach exhibit more flexibility and reduced reliance on the domain expert than the symbolic approach.
% Lastly, Amado et al. \cite{amado2022goal} created a data-driven value-based framework to solve GR problems via off-the-shelf value-based tabular RL algorithms: Q-Learning and Dyna-Q-Learning.
% 
% DL has been making significant inroads in recognizing patterns, such as pattern recognition and disease detection. In addition, its architecture is particularly suitable for processing sequential data, such as signals or text documents. Therefore, leveraging DL to solve GR problems is reasonable and includes a major flexibility improvement.
Both \citet{chiari2022goal} and \citet{maynard2019cost} use DL to solve GR problems. 
They propose neural network (NN) architectures that get observations as input and return the probability of reaching each goal. 
% However, these two papers are on opposing ends when looking at symbolic and data-driven representations as a spectrum. 
% Chiari et al. \shortcite{chiari2022goal} use Recursive NNs where the problem's input is a sequence of symbolic observations, and Maynard et al. \shortcite{maynard2019cost} use Convolutional NNs with image-based inputs. \frm{The type of neural network might not be that important here, perhaps we should focus on the data they use.}
% Since their neural network type is different, they also differ in the network's input.
% Maynard et al. \cite{maynard2019cost} used images because their GR framework is based on CNN.
% Nevertheless, Chiari et al. \cite{chiari2022goal} used parsed observations as the neural network input.
Although the DL approaches presented in these papers present major improvements, it also faces several difficulties.
\citet{chiari2022goal} discretises the continuous environment and hence suffers from the same discretisation issues discussed earlier. \citet{maynard2019cost} mitigate this by working directly with images of the domain. 
These approaches require extensive data collection, making them highly inflexible to changes and previously unseen states.  %adjusting to new domains is getting more difficult because it requires the data collected beforehand.
DRACO provides a compromise between these approaches: it uses unstructured input, but a distinct NN is allocated for each potential goal, so learning is focused and efficient. 
DRACO can recognize goals within all the available fluents of the domain as long as interacting with the environment (or a simulation) is possible, which is often a limitation in machine learning approaches for GR \cite{zhuo_shallow}. 

\section{Conclusion}\label{sec:conclusion}
    % !TEX root = ../main.tex
This paper provides a novel approach for end-to-end goal recognition in continuous domains. Its contributions include the DRACO algorithm, two distance metrics to compare observed trajectories to a policy, and a testbed of unstructured goal recognition environments and various problems. DRACO is shown to be more accurate, have higher confidence, and scale better than existing approaches, both in discrete and continuous domains.
% and when compared to symbolic and to learning-based approaches. 
%suitable for future approaches along the lines of DRACO. 
% While many GR approaches are symbolic and rely on manually crafted domain theories, DRACO automatically learns the environment dynamics from interactions without an expert specifying the transition model or its reward function. Additionally, 
DRACO has many advantages over symbolic and tabular RL approaches.
First, it works in both continuous and discrete domains without having the recognition process affected by the quality of a training set. 
Such approaches are limited by the states and goals within the dataset.
Second, its memory footprint and sample efficiency scale better, as earlier algorithms depend on the size of the state and action spaces.
Third, it doesn't require a domain-expert to provide the problem's formulation.
% This inability to scale is because tabular RL requires the agent to store each state-action pair in the domain and to visit such pairs numerous times before its value estimate converges, which is not viable in large state spaces. \frm{This conclusion should be obvious to any RL researcher}
% Our approach leverages state-of-the-art DRL techniques, so it is free of such limitations. Its memory consumption is constant and limited to the underlying NNs' size.
Finally, value-based RL techniques like Q-learning often fail continuous domains due to instability and poor convergence, leading to suboptimal policies.
% sometimes unstable, and often have poor convergence.
% which often leads to suboptimal policies.

%
% There are still many more open challenges related to recognition in continuous spaces that DRACO does not address: The current implementation works well in discrete action spaces, but it will not be feasible to compute $\pi_O$ when the action space is continuous.\frm{Surely this is not true, our action space is continuous is it not?} 
%

While DRACO is promising for robust goal recognition, it has three limitations:
% First, access to simulations: DRACO requires executing multiple instances to learn the goal-dependent policies. While this does not require DRACO to perfectly replicate the actor's policy, it should be able to produce policies that are different enough from one another. In the future, we aim to extend our work to use policies learned from Imitation Learning~\cite{HusseinEtAl2017} or Learning from Observation~\cite{zhu2020off} techniques. 
First, access to simulations: DRACO requires executing multiple instances to learn the goal-dependent policies. While this does not require DRACO to perfectly replicate the actor's policy, it should be able to produce policies that are different enough from one another. In the future, we aim to extend our work to use policies learned from Imitation Learning~\cite{HusseinEtAl2017} or Learning from Observation~\cite{zhu2020off} techniques. 
Second, number of goals: While DRACO's online inference is fast, it still needs to learn a new policy for each goal the actor might pursue. %This learning process is thus dependent on the number of goals the agent might be pursuing and can take a long time in domains with many or changing goals. 
Potential ways to overcome this are using universal value functions~\cite{schaul2015universal} or transfer learning techniques \cite{taylor2009transfer}.
Third, learning quality: More complex domains may require elaborate learning techniques. DRACO's PPO can be replaced by more sophisticated RL algorithms. 
    %, yet this investigation is beyond the scope of this paper. We note that additional ways exist to compute policy networks and $\pi_O$ to increase the algorithm's robustness to states from its training set. 
For example, image-processing techniques \cite{gedraite2011investigation} or policy acquisition techniques such as inverse RL \cite{ng2000algorithms} can fine-tune the policies. DRACO provides a consistent infrastructure through which such solutions can be implemented.

\bibliography{bibliography}

\end{document}